# Explainable Search and Discovery of Visual Cultural Heritage Collections with Multimodal Large Language Models

Taylor Arnold[1,*,†], Lauren Tilton[2,†]

[1]*Data Science & Linguistics, University of Richmond, U.S.A*
[2]*Rhetoric & Communication Studies, University of Richmond, U.S.A*

### Abstract
Many cultural institutions have made large digitized visual collections available online, often under permissible re-use licences. Creating interfaces for exploring and searching these collections is difficult, particularly in the absence of granular metadata. In this paper, we introduce a method for using state-of-the-art multimodal large language models (LLMs) to enable an open-ended, explainable search and discovery interface for visual collections. We show how our approach can create novel clustering and recommendation systems that avoid common pitfalls of methods based directly on visual embeddings. Of particular interest is the ability to offer concrete textual explanations of each recommendation without the need to preselect the features of interest. Together, these features can create a digital interface that is more open-ended and flexible while also being better suited to addressing privacy and ethical concerns. Through a case study using a collection of documentary photographs, we provide several metrics showing the efficacy and possibilities of our approach.

## 1. Overview

Numerous cultural organizations have digitized extensive visual collections and offered them online with licenses allowing flexible reuse [30]. These include national archives, major art museums such as the Rijksmuseum and the Louvre, and private institutions such as the Getty Museum and the Metropolitan Museum of Art [12, 15]. Third-party institutions, such as the MediaWiki project, the Google Art Project, and the Internet Archive, have also led efforts to produce visual corpora of cultural artifacts. These efforts correspond with movements within academic research to move beyond textual analysis toward visual and multimodal methods [6, 20, 32, 47]. Searching for keywords or individual works of art within (and across) these extensive collections according to existing structure metadata is relatively straightforward. But

CEUR Workshop Proceedings (CEUR-WS.org)

how do institutions help the public explore the breadth and depth of large visual collections *as visual archives* [28]?

It is quite an undertaking to build generous interfaces — what Whitelaw describes as "rich, browsable interfaces that reveal the scale and complexity of digital heritage collections" [48] — for visual cultural heritage collections [49]. Unlike digitized textual records, visual data does not come with the kinds of built-in search and similarity metrics that can be derived from word and n-gram counts [8]. One of two methods is typically used to overcome this difficulty. The first approach starts by selecting a set of pre-specified tags to describe each image. For example, we might tag images with their dominant colors, the number of people in the frame, or a list of the detected objects. These tags can be generated by manual tagging, crowd-sourced methods, or, more commonly, through the automatic application of computer vision algorithms [15, 17, 40]. Alternatively, abstract objects known as image embeddings can be used to associate each image with a sequence of numbers [14]. While each of the numbers is not individually meaningful, images with similar sequences of numbers will share common features [36]. Image embeddings are most commonly built using the internal representations of images within deep learning models built for object recognition [19, 35].

Distance metrics derived from either of these methods can be used to produce generous interfaces through the use of approaches such as cluster analysis and recommender systems. Building a generous interface from explicitly produced tags has the benefit of being able to explain the resulting structures. For example, suppose we tag images with the number of people present in the frame of the image. In that case, we can allow users to select images by the number of people in the image and expose this as an option in a faceted search interface. Using image embeddings, on the other hand, benefits by finding novel connections that can cut across existing categorization methods. However, relationships determined by image embeddings do not correspond to an immediately available description of why a set of images are associated with one another, making it challenging to use image embeddings for faceted search. Embedding-based connections also have the potential to produce connections between images that suggest or reinforce stereotypes and other implicit biases.

Recent advances in multimodal models offer the possibility of avoiding the choice between using fixed but explainable image annotations and flexible but abstract representations of visual data as embeddings. For example, Smits and Weavers recently showed the power of zero-shot learning for exploring historic collections [41]. They used the CLIP model to build classification algorithms for arbitrary tags without specifically training a model for a given category [37]. While the focus of their case studies was the analysis of specific subcategories (indoor/outdoor, family-based tags, and scene detection), they note the potential for a "new kind of bottom-up access to visual collections" through the application of multimodal models without the need for extensive manual annotations [41].

Over the past twelve months (mid-2023 through mid-2024), the integration of large language models (LLMs) and generative computer vision models has allowed for a radical increase in the capabilities of multimodal methods [1, 24, 50]. Current iterations of multimodal LLMs, such as Google's Gemini, OpenAI's GPT-4-Turbo and GTP-4o, and Apple's FERRT, allow users to submit an image and a textual prompt and receive a free-text response in return [53]. The results are not entirely free of errors [45], however the outputs have been shown to meet or exceed human annotations on a variety of sub-tasks, even without the need for customized fine-tuning [21, 46, 52]. Importantly, these multimodal LLMs far outperform previous methods for automatically captioning images and photographs [4, 25, 34, 38, 39]. This opens the possibilty for combining the benefit of explainable tag-based methods and unconstrained open-ended embedding-based methods for exploring large collections of digitized images.

In this paper, we present a general approach to using multimodal LLMs to search and discover vast image repositories. Our method first generates a set of automated captions for each image in the collection. Then, classical techniques from textual analysis are used to generate meaningful descriptions of the connections between images. We introduce a case study to evaluate how multimodal-based captions compare to those generated by visual embeddings. In the next section, we describe our approach in more detail and outline how we applied it to our selected collection. Then, in the following three sections, we offer qualitative and quantitative analyses of our approach by comparing it to image embedding-based techniques and showing the ability of the multimodal models to generate explainable connections. We conclude with a brief discussion showing how our approach can be extended and generalized.

## 2. Method

A typical workflow for working with an extensive collection of images is to use computer vision to either map each image into structured annotations (e.g., the number of people present) or to directly map the image into an abstract embedding space [2, 8]. Our method takes an alternative approach by using multimodal LLMs to produce rich captions as an intermediate surrogate. Text-based algorithms can then be applied to the resulting captions to produce similarity metrics, text-embeddings, and other summarizations. Conceptually, this can be described by the following flow of information:

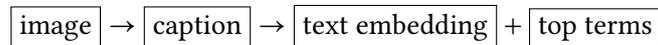

A significant amount of customization can be applied to this framework based on the needs of particular applications. The captions, for instance, could be exposed through a digital interface to allow for full-text search and increase accessibility. Or, if there is a concern that automatically generated captions may not be up to the metadata standards of the institution, they can be hidden from view and used only as the backend underlying a clustering analysis or recommender system. Different information can also be captured through prompt engineering and the choices of the models used.

In the remainder of this article, we show how this general approach can be applied to a collection of nearly sixteen thousand digitized documentary photographs created during the 1970s by the U.S. federal government as part of the Documerica project [5]. For our case-study, we used the OpenAI API for the caption creation and the text embedding. The total cost of producing the results in this paper were $287. The costs should scale linearly with the number of images and could be reduced by a factor of four or more by using the batch-based API and replacing intermediate steps with local techniques.

We started by taking each of the images in the collection and scaling them to have the largest dimension no greater than 1024 pixels and the smallest dimension no greater than 768. These sizes were chosen to optimize the price of the API request while being close to the maximum allowed size (testing suggested that smaller resolutions of the images produced much less accurate captions). We then made an API request using the GPT-4 Turbo model (version 2024-04-09) by submitting the image along with the query "Provide a detailed plain-text description of the objects, activities, people, background and/or composition of this photograph" [1]. The specific query was manually engineered after some trial-and-error using a test set of 25 images to get a complete description of different aspects of the image with a minimal amount of subjective commentary. We requested that the captions be a maximum of 500 tokens. Finally, we submitted the automatically generated captions to the OpenAI text embedding API (version 3). The API generated textual embeddings in a 3072-dimensional space. We then generated similarity scores between pairs of images using the cosine similarity between the textual embeddings. To provide a point of comparison, we also passed each image through the EfficentNet embedding using an open-source implementation [7, 44], generating a similar set of cosine similarity scores based only on the visual image.

Ultimately, we generated a rich caption and associated embedding for each image in the collection using a multimodal LLM. Using these embeddings, we were able to measure the distance between any pair of images. In the following section, we compared these with distances generated through an embedding generated directly from the image.

## 3. Qualitative Analysis and Global Structure

We ran the entire set of Documerica images through the method described in the previous section. Our analysis used the color-corrected images that account for the degradation of the online digitized photos [9]. On average, the automatically generated captions used 236 tokens (sd=47.1), corresponding to 197 words (sd=38.5). Two of the images had captions that could not fit within the 500 token limit specified in the search query. We also had two images that triggered the following warning message: "Your input image may contain content that is not allowed by our safety system," with no further output. One of the rejected images showed a scene with heavy fog. The other was a small object floating in a pool of a purple-colored liquid.

Two examples of the generated captions are shown in Fig. 1 and Fig. 2. The displayed captions are indicative of those found for all of the images. Captions typically start with a one-sentence overview of the scene shown in the photograph. Then, several sentences dive into specific objects, activities, and lighting conditions. When the model needs to make an inference based on partial information, the output often includes hedge phrases such as "appears to be" or "possibly". Over 80% of the captions include at least one of these phrases. Towards the end of the caption, the algorithm becomes more subjective, here giving comments about the "utilitarian" and "gloomy or overcast" ambiance of the photographs. Also, as seen in these examples, over half of the captions end with a summarizing statement that sums up what the algorithm believes to be the main message of the image. While most of text included in the captions appear to be both relevant and accurate, they are by no means foolproof. For example, the caption in Fig. 1 predicts that the worker is female, despite that not being at all clear from the image. The same caption also describes the objects in the foreground as "plastic", despite being made of glass.[1]

---

[1]The entire set of captions can be downloaded for further analysis from our website: https://distantviewing.org/

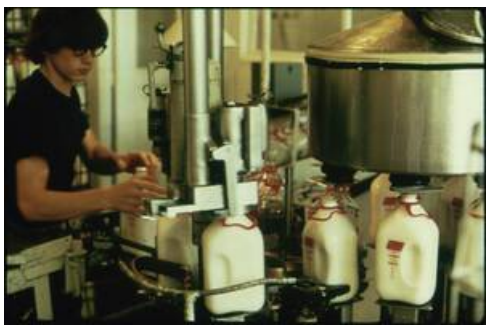

**Figure 1:** Documerica image (ID: 558353) with an automatically generated caption: "This photograph captures a scene inside an industrial or agricultural processing facility, possibly a dairy. The image focuses on a female worker engaged in operating machinery used for filling milk bottles. She is wearing glasses and a black top, and appears to be adjusting or handling the bottles or the machine. The worker is surrounded by a complex arrangement of stainless steel machinery and equipment. On the right side of the frame, there's a large metal vat or container, possibly holding milk or another liquid, with its lid partially visible. In the foreground, we see a series of white plastic milk bottles, each adorned with a red label. These bottles are lined up on a conveyor belt or a filling station, which seems to be part of a mechanized assembly line designed for packaging. The machinery includes various knobs, valves, metal structures, and hoses participating in the bottling process. The interior setting is lit with what appears to be fluorescent lighting, adding a utilitarian ambience to the workspace. The overall scene conveys a moment of daily operations wherein the worker is focused on ensuring the efficiency and correctness of the bottling process."

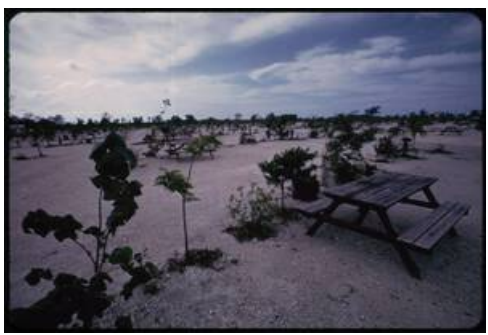

**Figure 2:** Documerica image (ID: 548671) with an automatically generated caption: "This photograph captures an open outdoor scene in a flat, barren park-like area with sparse vegetation. In the foreground, a wooden picnic table is situated prominently. The surface on which the table rests appears to be gravel, and it is located near a few scattered small bushes and very young trees. Beyond the picnic table, the landscape opens into a wider, desolate area that features numerous similar young trees and scattered bushes, all of which are relatively small and sparse, suggesting relatively young plantings in a possibly harsh environment. The soil looks rocky, further implying tough growing conditions. The sky above is cloudy, contributing to a somewhat gloomy or overcast ambiance. Overall, the setting appears to be a designed park area undergoing early stages of development, with the focus on newly planted vegetation in an otherwise empty, flat terrain."

---

downloads.

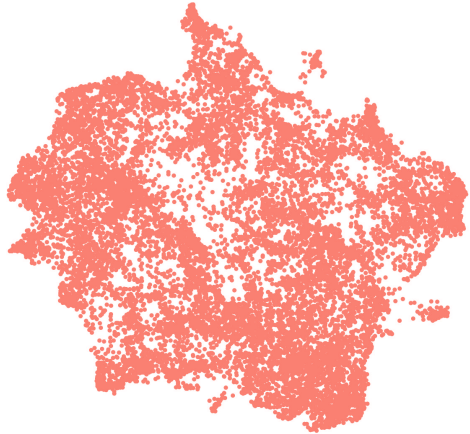 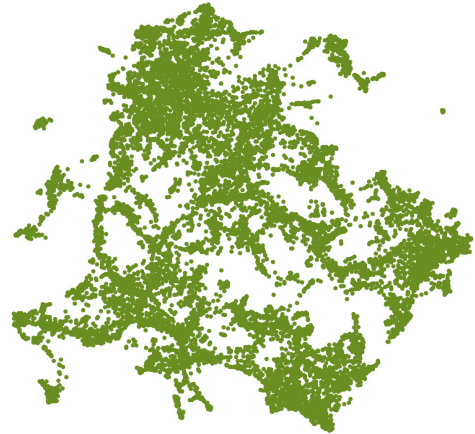

**Figure 3:** UMAP projection showing the structure of the text-based and multimodal embeddings. Each point corresponds to an image in the corpus; points near one another in the plot are also close together in the corresponding high-dimensional embedding spaces.

One way to understand the global structure of an embedding in a large vector space is to plot the output in a smaller dimension using dimensionality reduction techniques. A common choice for this is the UMAP dimensionality reduction projection. This algorithm tries to approximate the local structure of points in a high-dimensional space (here, the embedding space) in a lower-dimensional space [27]. Fig. 3 shows two-dimensional UMAP projections for the multimodal LLM and the embeddings directly derived from the visual input. The visual embedding displays larger continuous blocks of points, in contrast to the multimodal embedding, which has more corners, bridges, and distinct islands. These features indicate that the multimodal embedding identifies more distinct features. In the following section, we will investigate quantitative ways of measuring the differences between the two sets of recommendations.

## 4. Recommender System

How can we use the information in a set of embeddings to increase the access and discoverability of large collections? One common approach that has generally produced promising results across many collections is recommender systems [3, 13, 22, 29, 51]. Typically, recommender systems work by first allowing a user to pick an image (or providing one at random), and then suggesting a set of additional thumbnails of other related photos that may also be of interest. Clicking on a thumbnail shows a full version of the selected image and a new set of recommendations. Moving iteratively through a sequence of recommendations provides a unique, user-generated tour of a curated subset of a collection. At their best, the recommendations provide meaningful connections between images while avoiding getting users stuck within a small subset of the collection.

One way to build a recommender system is to provide recommendations based on the most similar images defined through similarity scores [23]. We already have two different sets of embeddings, those based on the captions and those from the visual embedding. We can create similar scores by computing the cosine similarity between the embedding vectors. These allow us to generate a set of $N$ recommendations for each image using the $N$ most similar images for any positive integer $N$ [33]. Building a recommendation system for a large set of images is an unsupervised learning task. There is no specific metric that we are trying to optimize for or ground truth that we are trying to reproduce. Therefore, we cannot reduce the summary between our two recommendation methods to a single number. Instead, we examine at several indirect measurements to compare the image-based and multimodal recommendation systems.

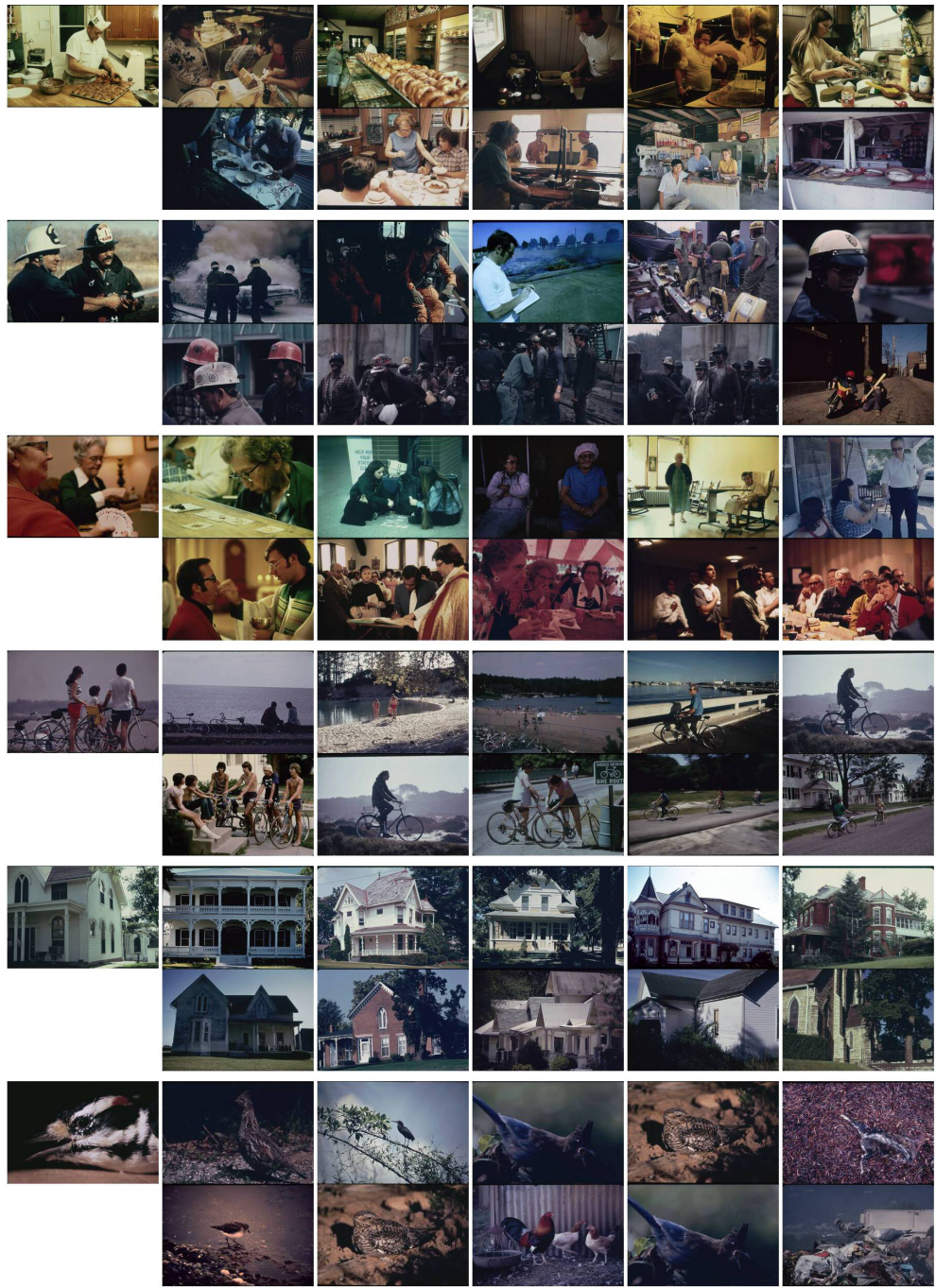

**Figure 4:** Six example images are shown on the left-hand side of the figure. Directly to the right of each of these images are the five closest recommendations based on the multimodal search. Below these recommendations are the five nearest recommendations based on the image-based recommendations.

**Table 1**
Percentage of symmetric recommendations—the recommendation to one image will include a back recommendation to the starting image—based on the total number of recommendations made.

| Num. Recommendations: | 1 | 5 | 10 | 15 | 20 | 25 |
|---|---|---|---|---|---|---|
| Image-Based | 22.4% | 24.9% | 26.2% | 27.3% | 28.2% | 28.9% |
| Multimodal-Based | 36.5% | 45.5% | 47.9% | 48.6% | 49.3% | 49.8% |

**Table 2**
Correspondence between the image-based and multimodal recommendations as a function of the number of recommendations. The table shows the average number of recommendations that are the same, the proportion of images that have no overlap between the two sets, and the proportion that have at most one overlap.

| Num. Recommendations | Avg. Overlap | No-Overlap | Overlap <= 1 |
|---|---|---|---|
| 1 | 0.11 | 88.8% | 100.0% |
| 5 | 0.59 | 58.7% | 86.7% |
| 10 | 1.24 | 38.5% | 67.0% |
| 15 | 1.97 | 26.3% | 51.4% |
| 20 | 2.74 | 19.1% | 39.5% |
| 25 | 3.57 | 13.9% | 30.7% |

Fig. 4 shows six sets of example recommendations. The photographs on the left-hand side show the starting images, with the five most similar multimodal recommendations on the top row and the five most similar text-based recommendations on the bottom row. Both sets of recommendations yield reasonably interesting results for these six selected images. The recommendations for the final image of a bird, for example, are very similar. However, the multimodal results generally offer recommendations that are both more precise and more diverse. For example, the fourth set starts with an image of three people with bicycles looking off into the distance. The visual recommendations only pick up on the bicycles, whereas the multimodal model also finds images with water in the background, including one image that does not even include bicycles. Similarly, for the fifth image of a house, the visual recommendations include rows of houses and a church; the multimodal recommendations only include single houses with similar architecture.

Another method of measuring the structure of the recommendations is to look at how often we have symmetric recommendations. In other words, if a specific image $A$ recommends an image *B*, we want to know how likely it is that image *B* will recommend back to image *A*. Having symmetric recommendations is generally a good feature because it indicates that the distance metric is meaningful and that we have a fairly uniform set of recommendations. Table 4 shows the proportion of symmetric recommendations for the two models based on the number of recommendations made. These proportions increase as the number of neighbors increase because there are more chances for them to map back into the original. In general, the image-based recommendations have a lower percentage of symmetric recommendations, with rates ranging from 22-29%, compared to the 36-50% rates of the multimodal recommendations. These correspond with the visualization shown in Fig. 3, which shows that the multimodal recommendations have many more tightly connected corners and clusters while still able ability to bridge between different parts of the corpus. These results indicate that the multimodal recommendations do a better job of finding tightly associated clusters. For this corpus, it finds these clusters without becoming too stuck in one particular part of the collection.

We can also directly compare how often the image-based and multimodal recommendations overlap. In Table 4, we show the proportion of the recommendations from each of the two methods that are the same as a function of the total number of neighbors. As we saw in the small set of examples in Fig. 4, there are a small number of overlapping recommendations. When using a recommendation size of ten, we average just over one matching recommendation. At the same time, the recommendations are not entirely disjoint. When we use a size of twenty-five, only 13.9% of images have no overlapping recommendations, with an average overlap of about 3.5. Based on these metrics, we see that the caption-based method produces noticeably different results from the image-based technique while preserving some similar structures.

## 5. Explainable Recommendations

A significant advantage of using captions as an intermediate step in the embeddings behind a recommender system is that we can use the captions to describe the rationale for associating two images. Specifically, once we have selected a fixed number of recommendations for each image, we can use the generated captions to produce a label that describes the set of relationships. Our approach was to first run the captions through an open-source NLP pipeline that performed tokenization, lemmatization, and part-of-speech tagging [43]. Then, we used log-likelihood scores to identify the nouns that most strongly differentiated the set of recommendations from the remainder of the corpus [42]. We selected the top five most strongly associated terms to label each set of recommendations.

We ran an experiment to test how well the generated labels correspond to the connections. First, we took a random set of 120 images and found the five closest recommendations for each from both methods. Then, we constructed the five most indicative terms for each set, creating separate sets for both recommendations. Then, we manually classified the proportion of recommendations that accurately corresponded to one of the terms. As a comparison baseline, we also took a random set of the generated terms from our set and counted the proportion of 500 randomly selected images that matched a given term. The results are shown in Table 5. The image-based tags matched at rates in the high 80s, whereas the multimodal tags matched in the high-to-mid 90s. These are all significantly higher than the randomly selected tags, indicating that the matches are not primarily a result of simply supplying generic terms. The biggest difference between the two recommendation systems is shown in the final column. Nearly 4% of the images fail to have any associated matching term. Only two of the multimodal-based terms match none of the terms. These results show that the top terms produced by the captions are relatively accurate and precise. While they can be used to add context to image-based recommendations, they perform noticeably better when applied to recommendations based on the captions' embeddings.

**Table 3**
Proportion of terms related to the referenced image based on the position of the term (top 5), with randomly selected images used as a reference point. The final column counts the proportion of recommendations that match at least one of the terms.

|  | Term # 1 | Term # 2 | Term # 3 | Term # 4 | Term # 5 | Any |
|---|---|---|---|---|---|---|
| Image-Based | 87.0% | 86.3% | 88.5% | 88.8 % | 84.5 % | 96.2% |
| Multimodal-Based | 97.8% | 96.1% | 95.8% | 94.3 % | 93.0 % | 99.6% |
| Baseline (Random) | 5.8% | 5.8% | 3.3% | 0.8 % | 5.0 % | · |

## 6. Clustering Analysis

Whereas recommender systems offer a way to explore similar images within a collection, how can the output of multimodal LLMs enable understanding the general themes within a collection of visual objects in the first place? Another application of caption text embeddings is to apply clustering algorithms that group together similar captions. Clustering has the advantage of being connected to the recommender system in the sense that images within a given cluster will tend to recommend other images within the same cluster. Also, similar to the approach in the previous section, we can use natural language processing techniques to find key terms that distinguish one cluster from all the others [42].

We applied a hierarchical clustering algorithm to the complete set of captions generated by our multimodal LLM [31]. The algorithm produced a set of 32 clusters, each tagged with the six terms that most distinguished it from all of the other clusters. These are shown in Table 6. The benefit of hierarchical clustering is that it allows us to generate a global structure on the clusters. Clusters in the table are ordered hierarchically so that clusters near each other on the table are more closely related than those farther away from one another. Those at either end of the table are the most unique and furthest away from the others.

Reading through the generated topics, starting at the top of Table 6, gives an understanding of the general structure of the Documerica collection. At the top are clusters associated with the detrimental effects of humans on the environment, such as pollution, waste, and junkyards. Then, we move to forms of transportation and into more productive transformations of the earth in the form of agriculture. We then transition into pure nature photos (cluster 15). Next, we see landscapes showing urban skylines and cityscapes. These move into other ways humans interact directly in their environment, such as hiking outdoors (cluster 28) and skiing (cluster 29). The final clusters correspond to particular shooting sets from parades, within laboratories,

**Table 4**
Clusters derived from a hierarchical cluster analysis using the text embeddings of the automatically generated captions. The labels are generated from log probabilities of the terms within each cluster. Clusters are ordered hierarchically so that clusters near each other on the table are more closely related than those farther away.

| ID | Cluster Description | Num. Photos |
| --- | --- | --- |
| 1 | landfill; environmental; waste; pollution; debris; garbage | 438 |
| 2 | old; decay; junkyard; car; destruction; scrapyard | 371 |
| 3 | helicopter; airport; urban; rainy; aircraft; cockpit | 140 |
| 4 | train; railway; track; railroad; station; maintenance | 220 |
| 5 | aerial; landscape; river; view; waterfall; natural | 1261 |
| 6 | industrial; facility; large; smoke; treatment; aerial | 1354 |
| 7 | outdoor; man; activity; group; people; picnic | 534 |
| 8 | man; elderly; portrait; older; technical; middle | 595 |
| 9 | man; elderly; conversation; candid; moment; couple | 240 |
| 10 | agricultural; rural; field; farm; crop; farming | 445 |
| 11 | flower; close; plant; up; cluster; delicate | 406 |
| 12 | sign; gas; store; market; billboard; advertisement | 514 |
| 13 | architectural; building; church; house; cemetery; story | 613 |
| 14 | car; parking; lot; vehicle; vintage; garage | 395 |
| 15 | bird; flight; close; surface; rock; deer | 660 |
| 16 | coastal; serene; beach; tranquil; lakeside; picturesque | 651 |
| 17 | landscape; forest; tree; sunset; dramatic; mountainous | 1311 |
| 18 | urban; cityscape; bridge; city; high; view | 611 |
| 19 | aerial; suburban; area; view; coastal; development | 172 |
| 20 | residential; house; suburban; street; story; neighborhood | 223 |
| 21 | highway; street; urban; busy; traffic; bustling | 628 |
| 22 | fishing; fish; underwater; net; water; coral | 375 |
| 23 | boat; sailboat; sailing; maritime; water; marina | 624 |
| 24 | beach; lakeside; day; activity; people; sunny | 343 |
| 25 | fountain; pool; public; park; urban; plaza | 87 |
| 26 | child; young; boy; girl; moment; playground | 351 |
| 27 | construction; industrial; site; mining; worker; machinery | 592 |
| 28 | woman; hiker; outdoor; young; individual; park | 420 |
| 29 | ski; resort; winter; snowy; snow; hockey | 89 |
| 30 | event; parade; street; public; vibrant; people | 544 |
| 31 | laboratory; room; woman; indoor; scientific; elderly | 335 |
| 32 | train; subway; station; interior; indoor; bus | 369 |

and photographs of trains and train stations.

The clusters generated here can be integrated into a digital platform that provides a generous interface for exploring the Documerica collection. Imagine, for example, a grid of thumbnails showing one image randomly selected from each cluster along with the associated keywords. Clicking on the thumbnail would create a page with a larger image version, archival metadata, and the recommender system described in the previous section. An option to return to the grid of clusters would be included prominently somewhere on the page. Such an interface would allow users to explore the expanse of the collection through each of the clusters while seeing the diversity within a cluster through the recommender system. Iteratively exploring the collection through these global and local connections would allow for a better understanding of the structure and overall message conveyed through the archive.

## 7. Conclusions

There are enormous possibilities for increasing modes of access, discovery, and analysis for visual collections through the automated generation of textual descriptions using multimodal LLMs. In this paper, we have introduced a general framework by which images can be converted into textual descriptions and text-based embeddings, opening them up to previously unavailable techniques. We applied an LLM, generated a certain kind of caption, and then used a recommender system and image clustering based on the text embedding of the caption. We showed how this approach could be applied to a collection of documentary photographs to produce an explainable recommender system and clustering-based descriptions of the themes within the collection. The present study is just one straightforward application of rich LLM-based multimodal methods. We expect to see a wide range of further applications of this general approach in the coming years, particularly as open-source models follow their usual pattern of catching up to the current state-of-the-art results currently attainable through closed, commercial systems [10, 26, 52].

We close with two specific extensions that highlight potential avenues of application for our framework. First, it is possible to add additional layers of safeguards to the recommendations, an important task when building interfaces to cultural heritage collections [11] This can be done through further prompt engineering or the filtering (or replacing) of terms before the embedding step. For example, we noticed that many terms in the captions, such as 'man' and 'girl', are gendered. As a result, the recommender system has a tendency to associate photos of people that it believes are the same gender, which in the case of people in the background is frequently based on inaccurate stereotypes [18]. Associations such as these can be mitigated, though never entirely avoided, by automated replacing gendered terms with neutral terms before running the text embedding. A second extension that can be implemented with the automatically generated captions would be offering an interface for a full-text search, allowing for new modes of accessibility [16]. Full-text search could be implemented to avoid the (not entirely correct) full captions themselves, or could expose these to end users along with a disclaimer about their autogenerated nature.